\documentclass[11pt]{article}

\usepackage[final]{acl}

\usepackage{times}
\usepackage{booktabs}
\usepackage{array}
\usepackage{xcolor}
\usepackage{geometry}
\usepackage{latexsym}
\usepackage{graphicx}
\usepackage{booktabs}
\usepackage{subcaption}
\usepackage{caption}
\usepackage{subcaption}
\usepackage{multirow}
\usepackage{multicol}

\title{\textsc{M-Help}: Using Social Media Data to Detect Mental Health\\ Help-Seeking Signals}

\author{MSVPJ Sathvik\thanks{This work is a result of equal contribution, with joint first authorship carried out during their undergraduate studies at IIIT Dharwad.} \\
  IIIT Dharwad \\
  Dharwad, India \\
\And
  Zuhair Hasan Shaik\footnotemark[1]\thanks{Corresponding author: zuhashaik12@gmail.com} \\
  IIIT Dharwad \\
  Dharwad, India \\
\And
  Vivek Gupta \\
  Arizona State University \\
  Arizona, USA
}

\begin{document}
\maketitle
\begin{abstract}

Mental health disorders are a global crisis. While various datasets exist for detecting such disorders, there remains a critical gap in identifying individuals actively seeking help. This paper introduces a novel dataset, \textsc{M-Help}, specifically designed to detect help-seeking behavior on social media. The dataset goes beyond traditional labels by identifying not only help-seeking activity but also specific mental health disorders and their underlying causes, such as relationship challenges or financial stressors. AI models trained on \textsc{M-Help} can address three key tasks: identifying help-seekers, diagnosing mental health conditions, and uncovering the root causes of issues.
\end{abstract}

\section{Introduction}
Mental health challenges such as anxiety, depression, and suicidal thoughts are global concerns. Despite growing awareness, stigma and the complexity of navigating mental health care systems create significant barriers to seeking help \cite{kalin2020critical}. These challenges are exacerbated by a lack of affordable services in underserved areas \cite{mcginty2023future} and limited guidance on accessing resources \cite{rifkin2023digital}. Although awareness campaigns reduce stigma, many remain unsure where to start, leading to delays or avoidance in seeking help \cite{wong2023psychological}.

Platforms like Reddit provide anonymous emotional support but lack the professional intervention needed for long-term recovery. While useful as a starting point, they cannot replace trained therapists or structured care. This underscores the need for solutions integrating online resources with accessible services like virtual therapy. A critical question arises: \textit{Can AI-powered platforms, leveraging advanced technologies like LLMs, effectively detect and support individuals facing mental or financial challenges?} To address this question, we propose a system that leverages social media data to identify individuals exhibiting help-seeking behavior. By developing a novel dataset \textsc{M-Help}\footnote{Data: https://huggingface.co/datasets/zuhashaik/M-Help} focused on detecting specific mental health conditions and their underlying causes, we aim to enable NGOs, government organizations, and other stakeholders to engage with at-risk individuals more effectively, offering personalized support and counseling.
\begin{table*}[t]
\setlength{\tabcolsep}{2.0pt}
\small
    \centering
    \begin{tabular}{p{11.0cm}|p{0.75cm}|p{1.75cm}|p{1.5cm}}
    \toprule[1.3pt]
        \textbf{Post} & \textbf{Help} & \textbf{Disorder}  & \textbf{Cause} \\
        \midrule[1.3pt]
        
         Nothing works and too depressed to help myself im 28, a man, i was bullied a decade ago for no go reason such as just being too different and socially awkward, got depressed, ive sought ...&Yes&MDD, PSTD, GAD&bullying and trauma\\
\midrule
 I'm not sure if I am being a jerk to myself or not and if I am, do I deserve it? I (44F) found out I have a tumor in my ...&Yes&MDD, GAD&Medication change and physical\\
\midrule
single mom as of recently ... need help!!!
let me start by saying i really dont know what or if to expect actual help. But when ...&Yes&ASD, MDD, GAD& financial hardships \\
        \bottomrule[1.3pt]
    \end{tabular}
    \label{tab:overview}
    \vspace{-0.5em}
    \caption{This table gives a overview of the Dataset \texttt{M-HELP}, more examples are given in the Appendix Table \ref{tab:dataset_example}.}
    \vspace{-1.5em}
\end{table*}

Prior research has laid a strong foundation in this space by introducing datasets and models for mental health disorder detection. For instance, \cite{mentalhelp,wexplainable,garg,trans, srivastava2025critical,srivastava2025sentimentguidedcommonsenseawareresponsegeneration,srivastava2025trustmodelingcounselingconversations} developed innovative datasets and machine learning models to detect mental health disorders. These systems leverage natural language processing (NLP) to analyze social media data, such as posts and interactions on platforms like Reddit and Twitter, to identify mental health issues through users' online behavior. Similarly, \cite{icwsm,kdd,jmir} created datasets and NLP models that summarize counseling sessions, emphasizing actionable feedback to improve therapy effectiveness and assist mental health professionals in planning interventions. Additionally, \cite{joyce2023explainable,arxiv1,adarsh2023fair} introduced explainable mental health detection datasets, focusing on identifying disorders and their underlying causes. By incorporating cause-related features, their work offers insights into users' mental states and circumstances.

Existing datasets and models have advanced mental health detection \cite{mazhar2025figurativecumcommonsense}, but their focus remains largely on disorder identification and counseling analysis and summarization \cite{srivastava2025sentimentguidedcommonsenseawareresponsegeneration,
srivastava2024knowledgeplanninglargelanguage}, leaving a critical gap in identifying and supporting individuals actively seeking help for mental and financial issues. Our work bridges this gap by introducing a novel dataset specifically designed to detect \textit{help-seeking behavior}, \textit{its underlying causes}, and \textit{mental health disorders} on social media, making it a holistic resource.

Our contributions are as follows:
\begin{enumerate}
\vspace{-0.25em}
\setlength\itemsep{-0.25em}
    \item \textbf{Novel Dataset for Help-Seeking Behavior:} We introduce the first dataset focused on detecting help-seeking behavior on social media.
    \item \textbf{Multi-Label Mental Health Disorder Classification:} We also incorporate multi-label classification data along with their underlying causes for the disorder.  
    \item \textbf{Evaluation of SoTA Models:} We evaluated advanced models such as GPT-4, Gemini-1.5, and Llama 3.1, proving their effectiveness in detecting help-seeking behavior.
\end{enumerate}

We will release the \textbf{M-Help} dataset and code upon acceptance to support psychologists, NGOs, and policymakers in connecting individuals with necessary resources.

\begin{table*}[ht]
\setlength{\tabcolsep}{2.75pt}
\small
    \centering
    \begin{tabular}{l|cc|cc|ccccccccc}
        \toprule
            Data &Avg words& Max words& (1) & (0) & MDD & OCD & BPD & GAD & SZD & ASD & AND & SUD & PTSD  \\ \midrule
        Train & 265& 3380 &648 &494& 676 & 117 & 446 & 601 & 120 & 78 & 25 & 206 & 285 \\
        Val   & 297& 2337 &191 &161&225 & 40  & 149 & 196 & 40  & 28 & 7  & 67  & 97 \\
        Test  & 274& 2795 &210&149&225 & 42  & 149 & 204 & 40  & 29 & 12 & 71  & 106 \\ \midrule
        Total & -&-&1049&804&1126 & 199 & 744 & 1001 & 200 & 135 & 44 & 344 & 488 \\
        \bottomrule
    \end{tabular}
    \vspace{-0.5em}
    \caption{\small This table present the entire statistics of the proposed dataset, \textsc{M-Help}. Label 1 represent the class 'Require Help' in the category Help-Seeking. And Avg words denotes the average number of words per post.}
    \label{stats}
    \vspace{-1.25em}
\end{table*}
\begin{figure*}[t]
    \centering
    \begin{subfigure}[b]{0.4\textwidth}
        \centering
        \includegraphics[width=\linewidth]{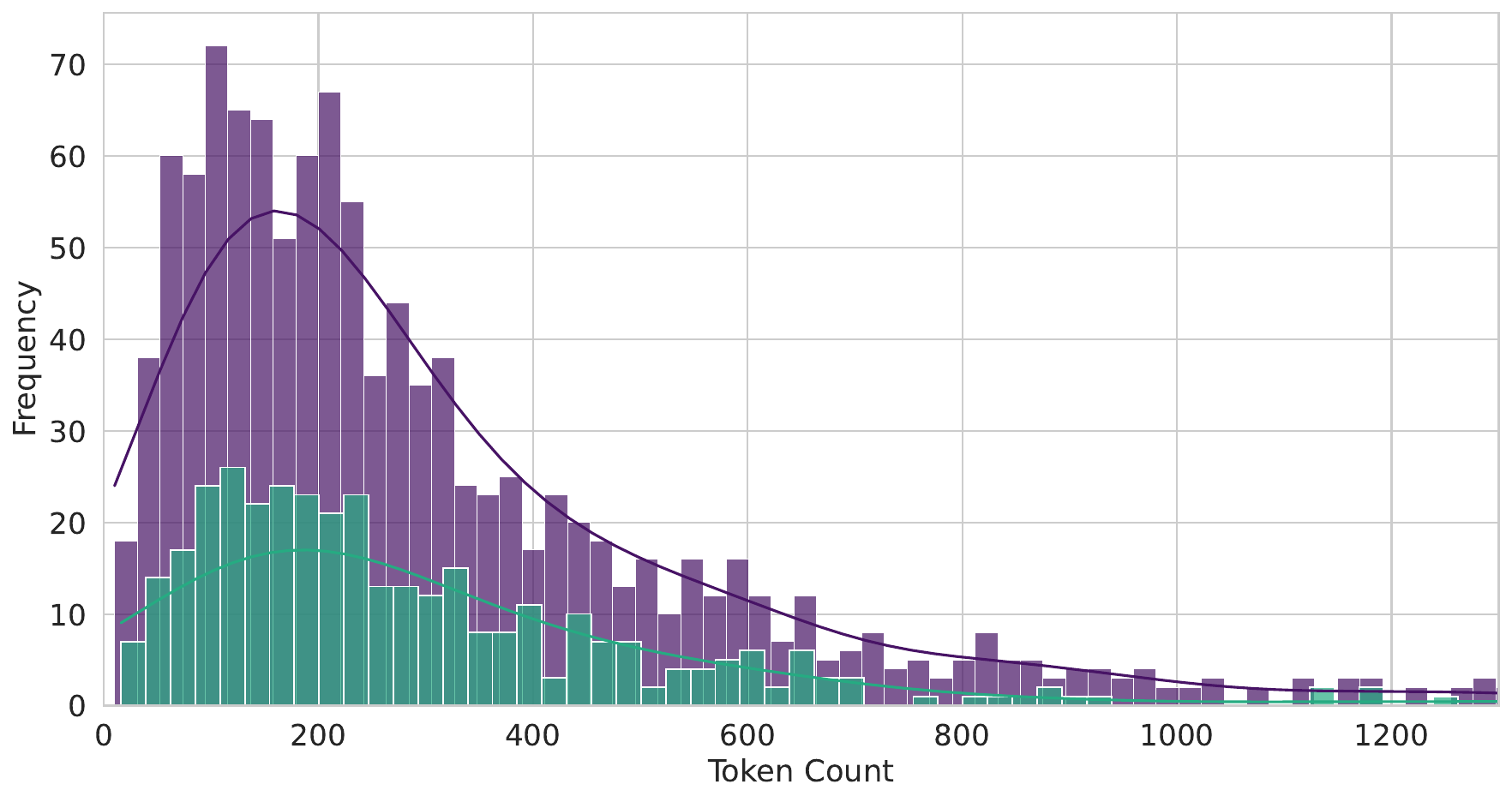}
    \caption{\small Token distribution in user posts and violet and green for train and test splits.}
        \label{fig:tokens}
    \end{subfigure}
    \hspace{1cm} 
    \begin{subfigure}[b]{0.35\textwidth}
        \centering
        \includegraphics[width=\linewidth]{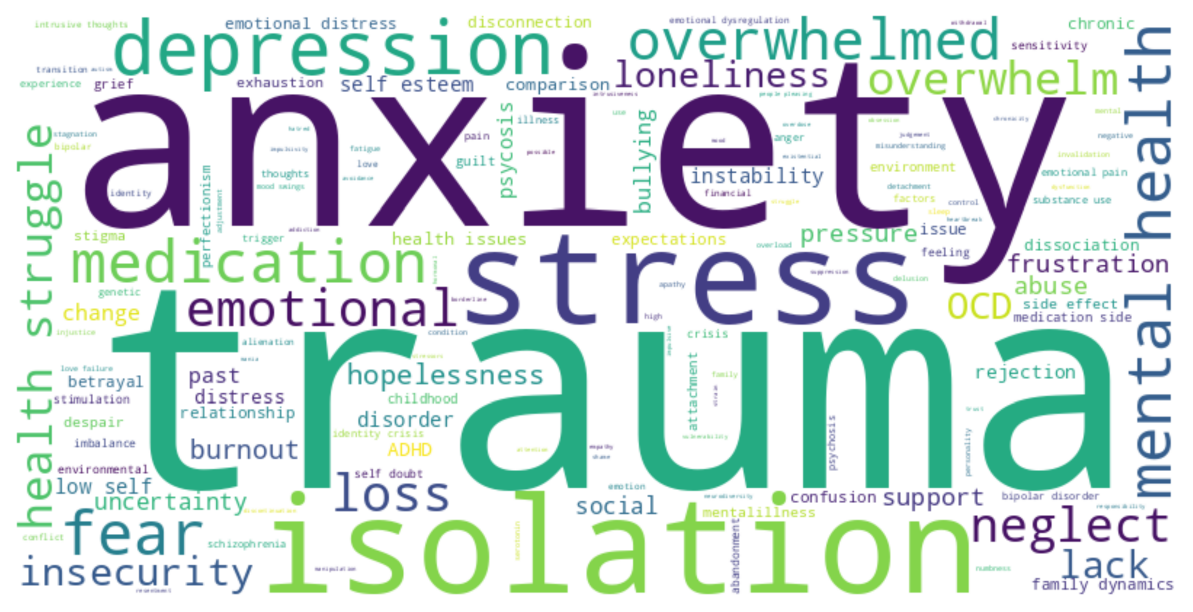}
    \caption{\small This figure shows the word cloud of the dataset for Cause.}
        \label{fig:wordcloud}
    \end{subfigure}
\vspace{-1.5em}
\end{figure*}

\begin{table*}[t]
\small
\centering
\setlength{\tabcolsep}{3.5pt}
\begin{tabular}{llccccccccccc}
\toprule
\multirow{2}{*}{Models Type } & \multirow{2}{*}{Models} & \multicolumn{4}{c}{Disorder Classification} & \multicolumn{3}{c}{Help-Seek Classification} \\
\cmidrule(lr){3-6} \cmidrule(lr){7-9}
& & Acc & Ma-F1 & W-F1 & Mi-F1 & Acc & Ma-F1 & W-F1 \\
\midrule
\multirow{3}{*}{\textbf{Encoder}} & BERT & 26.02 & 66.43 & 71.00 & 71.19 & 56.91 & 56.90 & 56.80 \\
& Mental-BERT & 25.75 & 66.32 & 71.89 & 72.14 & \textbf{61.25} & 58.80 & \textbf{60.18} \\
& RoBERTa & 29.81 & 65.84 & 72.62 & 73.08 & 59.89 & 55.90 & 57.73\\
\midrule
\multirow{3}{*}{\textbf{+ Long Context}} & Longformer-4096 & 23.04 & 50.58 & 63.59 & 65.19 & 57.18 & 57.14 & 57.32 \\
& AIMH/mental-longformer-base-4096 & 22.49 & 55.19 & 65.80 & 67.08 & 58.27 & 58.07 & 58.47 \\
& XLM-roberta-longformer-4096 & 25.47 & 59.92 & 70.44 & 71.00 & 59.08 & 57.64 & 58.72 \\
\midrule
\multirow{3}{*}{\textbf{Encoder-Decoder}} & BART & \textbf{35.23} & 67.54 & 75.11 & 76.07 & 57.72 & 56.19 & 57.32 \\
& Mental-BART & 32.79 & 72.93 & 75.42 & 75.55 & 59.89 & \textbf{58.98} & 59.83 \\
& T5 & 30.89 & 64.12 & 71.62 & 73.46 & 55.01 & 53.83 & 54.85 \\
\midrule
\multirow{2}{*}{\textbf{Decoder-Only}} & Phi-3.5 & 14.91 & 20.18 & 40.66 & 51.39 & 56.37 & 55.18 & 56.19 \\
& Llama-3.1 & 15.99 & 23.45 & 43.87 & 52.83 & 53.66 & 53.60 & 53.50 \\
& Mental-Llama & 17.07&	27.42	&49.56	&57.24	&58.81	&58.39	&58.87 \\
\midrule
\multirow{2}{*}{\textbf{API (Closed Source)}} & GPT-4o & 30.16 & 38.03 & 41.82 & 56.16 & 57.18 & 36.96 &41.89\\
& Gemini-1.5 & 32.52 & \textbf{77.95} & 
\textbf{76.59} & \textbf{77.90} & 57.99 & 38.98 & 43.68 \\
\bottomrule
\end{tabular}
\vspace{-0.75em}
\caption{Performance comparison of models across disorder and help-seek classifications, separated by model architecture categories. Values are represented as percentages.}
\label{tab:merged}
\vspace{-1.25em}
\end{table*}

\section{Construction of M-Help}

\subsection{Data Collection and Annotation}

To develop the dataset on mental health help-seeking behavior on social media, we focused on Reddit (A \ref{reddit}) communities where users discuss mental health challenges. After analyzing several subreddits, we identified \texttt{r/MentalHealthSupport} and \texttt{r/mentalillness} as the most relevant. Posts were extracted featuring users' personal mental health experiences, ensuring the dataset includes first-hand accounts. The dataset was annotated by a team of three mental health professionals, each with a minimum of three years of expertise in mental health therapy and domain knowledge in the field. This professional background ensured that the annotations were not only accurate but also informed by a deep understanding of mental health terminologies. The annotators developed detailed guidelines to ensure consistency and quality throughout the process. They co-validated their work, adhering to these guidelines. Due to the complexity of the user posts, the annotation took about eight weeks, allowing thorough review and labeling of help-seeking behavior, mental health disorders, and their causes.

\subsection{Categories and Criteria}
\label{main:cls}
\paragraph{a. Help-Seeking.}
A post is classified as "Help-Seeking" if the user is actively or indirectly seeking assistance related to mental health. While some users may directly ask for help, others, due to the anonymity of social media, may express their struggles in a more subtle manner. These users may not explicitly state their need for help, but the underlying request is conveyed through hidden meanings or indirect language in their posts. For instance, posts like "I feel lost and don't know what to do" or "I just can't seem to cope anymore" may indicate a need for support, even though the user does not explicitly ask for help. Our goal is to identify these underlying expressions of help-seeking behavior. We achieved a Fleiss' kappa score \cite{fleiss1971measuring} of 0.8355, which falls under Almost Perfect Agreement.

\paragraph{b. Mental Health Disorders.}
The disorders in this category are based on the criteria set by the \textit{Diagnostic and Statistical Manual of Mental Disorders DSM-5} \cite{regier2013dsm} and the \textit{International Classification of Diseases ICD-10} \cite{manchikanti2011necessity}. We selected 9 mental health disorders that cover a broad spectrum of mental health issues. The disorders included in our dataset are as follows:

\vspace{-0.25em}
\noindent - \textbf{Autism Spectrum Disorder (ASD)}: A developmental disorder affecting social interaction, communication, and behavior. \\
- \textbf{Schizophrenia Disorder (SZD)}: A severe mental disorder characterized by distorted thinking, hallucinations, and disorganized speech/behavior. \\
- \textbf{Bipolar Disorder (BPD)}: A mood disorder marked by extreme mood swings, from manic episodes to depressive states. \\
- \textbf{Major Depressive Disorder (MDD)}: A mood disorder characterized by persistent sadness, lack of interest in activities, and hopelessness feelings. \\
- \textbf{Generalized Anxiety Disorder (GAD)}: A condition involving excessive, uncontrollable worry about everyday situations. \\
- \textbf{Obsessive-Compulsive Disorder (OCD)}: A mental disorder involving persistent, unwanted thoughts and repetitive behaviors or rituals. \\
- \textbf{Post-Traumatic Stress Disorder (PTSD)}: A condition that occurs after a traumatic event, causing flashbacks, nightmares, and severe anxiety. \\
- \textbf{Anorexia Nervosa Disorder (AND)}: An eating disorder characterized by an intense fear of gaining weight, leading to restrictive eating and a distorted body image. \\
- \textbf{Substance Use Disorder (SUD)}: A disorder involving the harmful use or dependency on substances such as alcohol or drugs.
\vspace{-0.25em}

These disorders are multi-label, meaning a post can mention more than one disorder, such as both anxiety and depression, and be labeled under both Generalized Anxiety Disorder (GAD) and Major Depressive Disorder (MDD).

\vspace{-0.5em}
\paragraph{c. Cause for Help-Seeking.} This label (cause mention) is used when a post identifies specific causes or events that have triggered the user’s mental health struggles. Such posts often describe how particular life events, situations, or conditions have led to or exacerbated their mental health issues often using phrases like "because of" or "due to," or critical terms like suicide or "wanna die," e.g., "My anxiety started after a traumatic experience". This category helps to identify not just the mental health condition but also the underlying factors that may be contributing to the individual’s struggles, providing a deeper context for their help-seeking behavior. A word cloud for this category reveals key terms such as 'Anxiety,' 'Trauma,' 'Isolation,' and 'Stress,' among others. The detailed statistics of our proposed dataset \textsc{M-Help} is illustrated in the Table \ref{stats}.
We provide further details about our dataset, \texttt{M-HELP}, in the Appendix section \ref{a:dataset}.

\section{Experimental Details}

\paragraph{Baselines.} We evaluated a range of baseline models for Disorder Classification and Help-Seeking Classification, covering encoder-only, long-context, encoder-decoder, decoder-only, and closed-source architectures. Performance metrics, including Accuracy, Macro F1, Weighted F1, and Micro F1 for multi-label classification, are presented in Table \ref{tab:merged}. Given the length of user posts, traditional encoder models with limited context lengths (e.g., BERT, RoBERTa) face challenges, as shown in Figure \ref{fig:tokens} and further illustrated in Appendix \ref{app_tokens}. To address this, we employed long-context models such as Longformer and leveraged mental health-specific adaptations fine-tuned on domain datasets to enhance performance. Additional experimental details, including training and hyperparameters, are discussed in Appendix \ref{app:exp}.

\paragraph{Results and Analysis.} Table \ref{tab:merged} summarizes the performance of various models on disorder and help-seek classification tasks, comparing encoder-based, long-context, encoder-decoder, and decoder-only architectures.

\textbf{Disorder Classification:}
For disorder classification, Gemini-1.5 achieves the highest weighted F1 score (76.59\%), followed by Mental-BART with (75.42\%). Gemini-1.5 demonstrates strong performance in this task, but the Mental-BART variant, fine-tuned on mental health data, performs comparably. This suggests that Mental-BART could serve as a strong open-source alternative for impactful research in this domain.
Whereas larger decoder-only models like Phi-3.5, Llama-3.1, and Mental-Llama still struggle in this task, achieving significantly lower weighted F1 scores. Their inability to match the performance of encoder-decoder or closed-source models.

\textbf{Help-Seek Classification:}
For help-seek classification, Mental-BERT achieves the highest weighted F1 score (60.18\%), followed by Mental-BART with (59.83\%). This highlights the effectiveness of mental health-specific fine-tuning in improving classification performance.


\textbf{Key Observations:} Google, a well-known leader in health-related research, has developed models such as Gemini-1.5 that demonstrate significant improvements in performance. However, despite these advancements, they still struggle to effectively capture help-seeking signals. Models fine-tuned on mental health data, such as Mental-BERT, Mental-BART, and Mental-LLaMA, have shown significant progress in classifying both help-seeking behavior and mental health disorders. However, the current scores remain insufficient to produce high-quality mental health systems, underscoring the need for continued research and collaboration within the community to advance this field and build more reliable mental health detection models. \textsc{M-Help} provides a critical benchmark, enabling further research and the development of improved models for mental health detection on social media, where advancing research is essential to address the growing need for mental health support.

\section{Conclusion and Future work}  

In this paper, we introduce \textsc{M-Help}, a novel dataset designed to identify help-seeking posts related to mental health issues and their underlying causes. Unlike traditional datasets, \textsc{M-Help} uniquely focuses on capturing active help-seeking behavior, addressing a critical gap in mental health research. We benchmark state-of-the-art (SoTA) models, including GPT-4o, LLaMA-3.1, and other leading transformer-based architectures, to evaluate the dataset’s effectiveness. Additionally, we propose a multi-label disorder classification framework that not only identifies mental health disorders but also determines the underlying causes of distress. This work fills a crucial void in the mental health research community, providing a structured benchmark for identifying distress cues where timely intervention is needed. By expanding \textsc{M-Help} to support multiple languages and incorporating advanced detection techniques, future research can significantly enhance its effectiveness, fostering a more inclusive and impactful global mental health support system.



\section*{Limitations}

The main strength of the dataset is its real-time applicability and uniqueness. However, it lacks multilingual coverage. Although we acknowledge the importance of incorporating data in various languages to capture a wider range of perspectives, the absence of data in those languages within the subreddits led us to focus solely on English. As a result, the dataset only contains content in English.

One other limitation of the dataset is its exclusive focus on the Reddit platform. The data was collected solely from Reddit, an anonymous platform. Although data from Twitter and other social media platforms could have been scraped, doing so would violate recent ethical guidelines, and these platforms are not anonymous. As a result, the dataset is limited to Reddit.

\section*{Ethics Statement}

We, the authors of this work, affirm that our research adheres to the highest ethical standards in both research and publication. Throughout the course of our study, we have carefully considered and addressed various ethical aspects to ensure the responsible and fair application of computational linguistics methodologies. The findings presented in this paper are consistent with the results of our experiments. However, we acknowledge that some level of stochasticity is inherent in the use of black-box large language models. We mitigate this by maintaining a fixed temperature during our experiments. We also provide thorough descriptions of the annotations, dataset splits, models, and prompting methods used, ensuring that others can reproduce our work accurately. For grammar correction, we utilized AI-based writing assistants, and for coding tasks, we employed Copilot. It is important to emphasize that the development of our ideas and the execution of the research were entirely independent of AI assistance.

\bibliography{refs}

\newpage
\appendix
\section*{Appendix}
\section{\bf Training and computation Setup.}
\label{app:exp}
For the baseline training, all models were initialized from Hugging Face and trained for 10 epochs using three NVIDIA Tesla V100 GPUs. The proposed model was also trained for 10 epochs, with each epoch taking approximately 20 minutes. Inference time for classifying an intensity or target class was around 5 seconds. All training and inference operations were conducted on the same three NVIDIA Tesla V100 GPUs.

We employed the AdamW optimizer with a learning rate of $5 \times 10^{-5}$, $\beta_1 = 0.9$, and $\beta_2 = 0.999$. To mitigate overfitting, we applied a dropout rate of 0.2 and a weight decay of $1 \times 10^{-2}$. For single-label classification tasks, Cross Entropy Loss was used, while Binary Cross Entropy Loss was applied in multi-label scenarios. Model performance was assessed using Micro, Macro, and Weighted F1 Scores to ensure a comprehensive evaluation across different class distributions and task complexities.

The learning rate scheduler was set to constant, with a batch size of 16. The RMS Norm Epsilon was set to $1 \times 10^{-5}$, and the Adam Epsilon to $1 \times 10^{-8}$. Maximum sequence length was capped at 512 tokens. To stabilize training, gradient clipping was applied with a threshold of 1.0.

These hyperparameters were carefully tuned to optimize model performance while balancing computational efficiency. Table~\ref{tab:hyperparameters} details the hyperparameters used in our experiments, including additional parameters such as warmup steps and specific optimizer settings. This configuration allowed us to effectively train and evaluate our models across various mental health meme classification tasks, supporting both single-label and multi-label scenarios while ensuring robust performance through diverse F1 score calculations.

\begin{table}[h]
\centering
\begin{tabular}{ll}
\hline
\textbf{Hyperparameter} & \textbf{Value} \\
\hline
Learning Rate (lr)      & $5 \times 10^{-5}$ \\
Adam Beta1              & 0.9                 \\
Adam Beta2              & 0.999               \\
Adam Epsilon            & $1 \times 10^{-8}$  \\
RMS Norm Epsilon        & $1 \times 10^{-5}$  \\
Dropout                 & 0.2                 \\
Batch Size              & 16                  \\
Learning Rate Scheduler & Constant            \\
\hline
\textbf{Parameters $>$ 2 Billion} & \\
Lora Rank (r)          & 16                  \\
Lora Alpha ($\alpha$)  & 8                   \\
Target Modules         & $W_q, W_k, W_v, W_o$ \\
\hline
\end{tabular}
\caption{Hyperparameters used in the experiment.}
\label{tab:hyperparameters}
\end{table}

\section{More details on our proposed dataset M-HELP.}
\label{a:dataset}
\subsection{Help-Seeking Task Definition}
As described in section \ref{main:cls} of our paper, the help-seeking classification is framed as a binary task:

\paragraph{Definition}
A post is classified as \textit{Help-Seeking} if the user is actively or indirectly seeking assistance related to mental health. While some users explicitly ask for help, others express their struggles in a more subtle manner. Examples include:\\
\textbf{Direct:} "I need help coping with anxiety."\\
\textbf{Indirect:} "I feel lost and don’t know what to do."\\

The objective is to identify both direct and indirect expressions of help-seeking behavior.

\subsection{Data Source and Compliance}
\label{reddit}
Reddit, as an open-source platform, allows individuals to express themselves freely. Data was collected using Reddit’s API in full compliance with its terms of service. Unlike other social media platforms (e.g., Twitter/X, Instagram), which impose access restrictions, Reddit provides an ethical avenue for research without compromising user privacy.

\subsection{Annotation Process}
The dataset was annotated by licensed medical therapists and practitioners. These professionals conducted discussions and assigned labels based on established psychological frameworks, specifically the \textit{Diagnostic and Statistical Manual of Mental Disorders (DSM-5)} and the \textit{International Classification of Diseases (ICD-10)}.

\section{LLMs are confused between demotivating and mental health disorder}
Through our analysis and benchmarking, we've found that LLMs often confuse demotivation with mental health issues, even when the text clearly points to one or the other.

\textit{For example:}

\textit{how to fix myself} 

\textit{hi I'm 22 yo I'm struggling to find motivation to live or just how to get through this. I am an introvert, I have a big problem with establishing contact and communicate with people. I have spent most of my life at the computer. I'm struggling to find a job. I don't rly have any friends and I know I can't talk to my family about my feelings.I have 4 adult siblings and I'm the youngest.}

\textit{Because I couldn't find a job i went to college and last week I made decision to quit ( field of study didn't suit me) . I feel like I wasted so much time and going there was a big mistake. I know my family is mad at me for that and I keep disappointing them. For now I only live with my mom and we have just enough money to go from month to month. i am a liability . So I rly need to find a job. I don't know what to do in live, I never had any life goals, for now I just don't want my family to be mad at me.} 

\textit{I rly hate myself why am I like that. I know I shouldn't compare myself to others but other people have jobs, go to college have relationships and I'm struggling with simple things. when I think about future I feel like my chest is about to explode. I remember my summer job at the warehouse I worked there maybe less than a month because of broken hand. Work itself wasn't bad but interactions with people were horrible. I couldn't bring myself to enjoy any of these conversations. Lots of fake smiles. I hated it there. Idk if I would be able to handle another job like this. }

\textit{Lately I'm been starting to think more seriously about disappearing. Nothings is rly keeping me here. And I just don't know how long I can handle it I feel like I'm about to crush.}

Actual label: 0

Prediction: yes

Although the user explicitly stated that they needed help with motivation, most of the LLMs interpreted it as a request for mental health support.

\renewcommand{\arraystretch}{1.2}
\setlength{\tabcolsep}{8pt}

\begin{table*}
\centering
\footnotesize
\resizebox{\textwidth}{!}{%
\begin{tabular}{p{58em}} 
\toprule
\textbf{ID: ID-2} \\
\\
\textbf{Text:} I (26F) have struggled most of my life with mental health issues. A lot of it stemmed from a difficult childhood and as a result, I’ve been struggling with anxiety and depression since 14.
Throughout my adult years, I’ve struggled with unhealthy coping mechanisms like drinking, kissing/sleeping with and dating men who weren’t nice etc. Last year in therapy, I made a breakthrough with dealing with some of the family trauma and though things felt a bit better, I was still struggling a lot. I got with my boyfriend during this time and he was supportive throughout and we had the perfect relationship. I decided to quit my job (which was a big contributor to my unhappiness) and go travelling for 6 weeks by myself. Just before I left, my boyfriend fell into a depressive period which I struggled to support him with. While I was travelling, I realised the deep rooted reason I was always struggling was because I am very insecure and have pretty much no self esteem. While I was trying to deal with this, I found it increasingly harder to be there for my boyfriend as well and we decided to break up as we both needed time to work on ourselves. It was awful as we still very much love each other. The break up was a turning point for me to start taking the actionable steps to making my life better as I realised it didn’t matter what my circumstances were, I needed to heal for myself.
I reached out 6 weeks later saying I’d reached a place that I could support him better and wanted to try again but he said he wasn’t in the right place mentally (which is where I was at when we broke up). He’s also still hurt bc v soon after we broke up, I got really drunk and kissed someone which he was devastated about but was me turning to old self destructive habits which I regret and have worked to make some deep fundamental changes to not act like that again. 
It’s been almost a month since I feel like I went through a 2nd break up with him and since then I’ve been hit with a million realisations about myself \& my entire foundations of who I am and how I deal with life completely fell apart. I know this is a good thing as it means I can rebuild stronger but I am now in a place of feeling hopeless as I don’t know what to think or feel anymore. I can feel new and much healthier thought patterns forming and I’ve stopped going back to old behaviours but day to day I am really struggling and can’t stop crying. Everything feels pointless because I’m so scared of feeling like this forever. It’s like the  old fighting the new. I’ve heard things get worse before they get better but I’m scared as it feels like they keep getting worse after they feel slightly better. I go to therapy but it’s the day to day where I feel lost and I can’t focus on the present. I’m constantly stressing that I’ll forever feel like this and the thoughts don’t go away even when I’m trying to distract myself. I miss my ex so much, I hate that I hurt him and I hate that I couldn’t be there for him in the way I feel I could be now and I feel immense guilt for how I handled everything. It feels like my life has completely flipped - I had to move back home, away from most my friends, I came back from travelling with no money, i decided to come off my SSRIs after six years (i felt in a better place to deal with things), I have started a new job a month ago and I am heartbroken. I have to keep going to the toilets at work to cry throughout the day. 
If anyone has been in a similar situation or can give advice please do as I’m struggling to carry on. Will it get better? And if so, when did you start to feel things were getting easier? I don’t know how much more of this I can take. \\
\textbf{Help:} 1 \\
\textbf{Cause:} love failure \\
\textbf{MH Condition:} ['BPD', 'GAD', 'MDD', 'SUD'] \\
\midrule
\textbf{ID: ID-4} \\
\\  
\textbf{Text:} I can’t take this shit anymore I just can’t do it ever since I been in my 20s life has been nothing but hell I can’t do this anymore I want it all to stop it’s not worth the heart breaks it’s not worth crying yourself to sleep everyday none of it is worth living i don’t know what I did to deserve this but I’m so sorry for whatever I did I’m so sorry just please make it stop I can’t take it anymore \\
\textbf{Help:} 1 \\
\textbf{Cause:} trauma and unresolved pain \\
\textbf{MH Condition:} ['BPD', 'MDD'] \\
\midrule
\textbf{ID: ID-5} \\
\\
\textbf{Text:} I have diagnosed ADHD, and of course you'd think it would come with a little bit of executive dysfunction, but not to the point in which it has gotten when it comes to taking care of my teeth. I also used to be bad at washing my face but I found a way to motivate myself to do so, but my ability to bring myself to care about my dental health is, to be quite honest, sad. I used to have braces years ago, and I think that's what really kickstarted my executive dysfunction when it comes to even little things like brushing my teeth or wearing my retainer at night. I know that what I am doing could actually become something worse like serious gingivitis and I do brush my teeth, but never nearly as often as I know I should. My brain will not let me brush my teeth on a regular basis, it refuses to allow me the motivation to do so, to the point where I'm honestly having anxiety about it. But I can't just do what I did with my struggles around washing my face and just brush them in the shower. I don't know what to do regarding my brains apparent love for making me unmotivated in my dental care, and I really need help with finding a solution. \\
\textbf{Help:} 1 \\
\textbf{Cause:} executive dysfunction \\
\textbf{MH Condition:} [GAD] \\
\bottomrule
\end{tabular}%
}
\caption{Dataset examples illustrating mental health issues, causes, and conditions.}
\label{tab:dataset_example}
\end{table*}
\section{Token Distribution}

\begin{figure*}[h]
    \centering
    \includegraphics[width=0.8\textwidth]{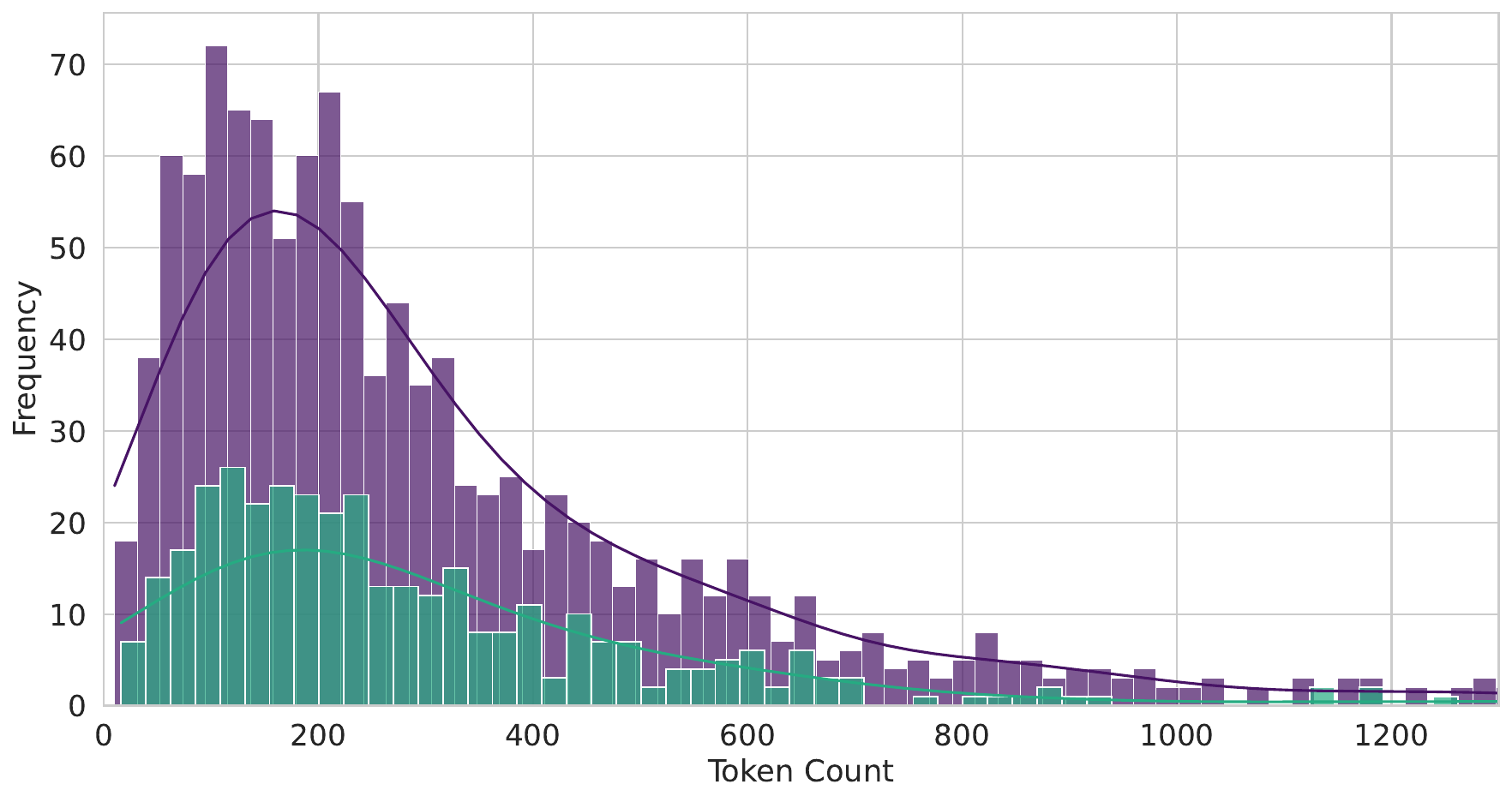}
    \caption{Token distribution for \textbf{Mental-BERT-base-uncased} in user posts. Violet and green represent train and test splits, respectively.}
    \label{fig:tokens-mental-bert}
\end{figure*}

\begin{figure*}[h]
    \centering
    \includegraphics[width=0.8\textwidth]{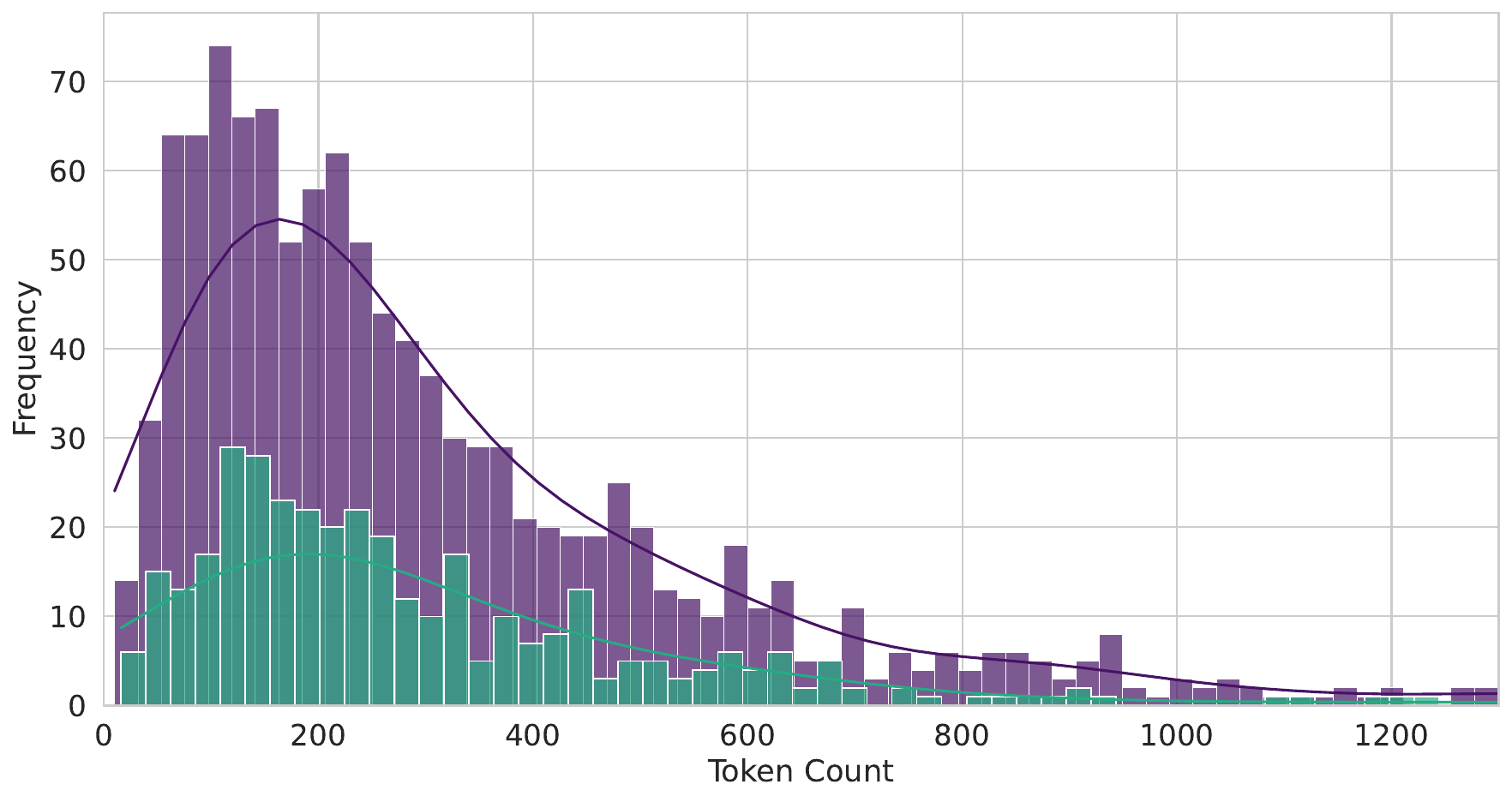}
    \caption{Token distribution for \textbf{RoBERTa-base} in user posts. Violet and green represent train and test splits, respectively.}
    \label{fig:tokens-roberta-base}
\end{figure*}

\begin{figure*}[h]
    \centering
    \includegraphics[width=0.8\textwidth]{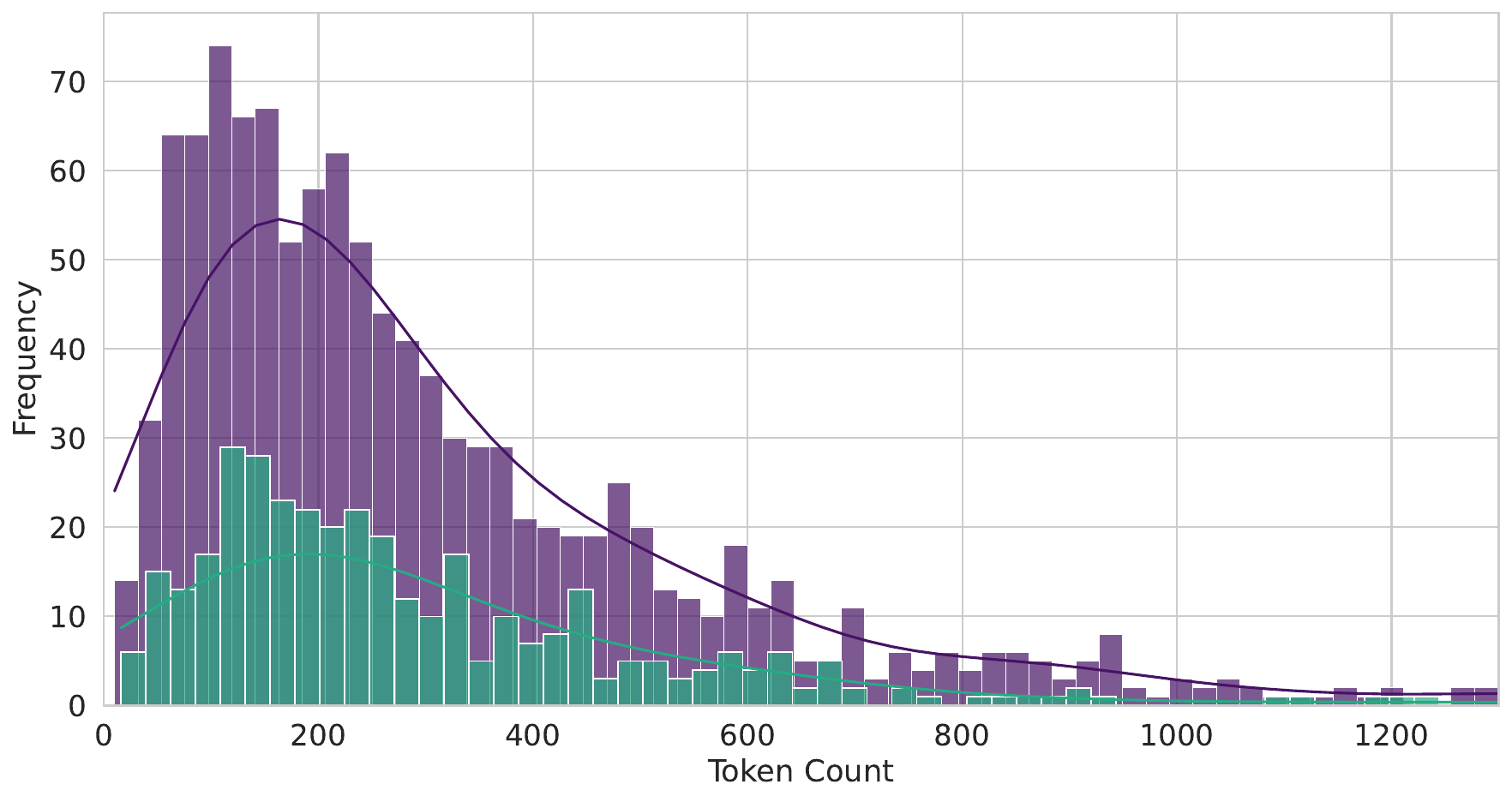}
    \caption{Token distribution for \textbf{Longformer-base-4096} in user posts. Violet and green represent train and test splits, respectively.}
    \label{fig:tokens-longformer}
\end{figure*}

\begin{figure*}[h]
    \centering
    \includegraphics[width=0.8\textwidth]{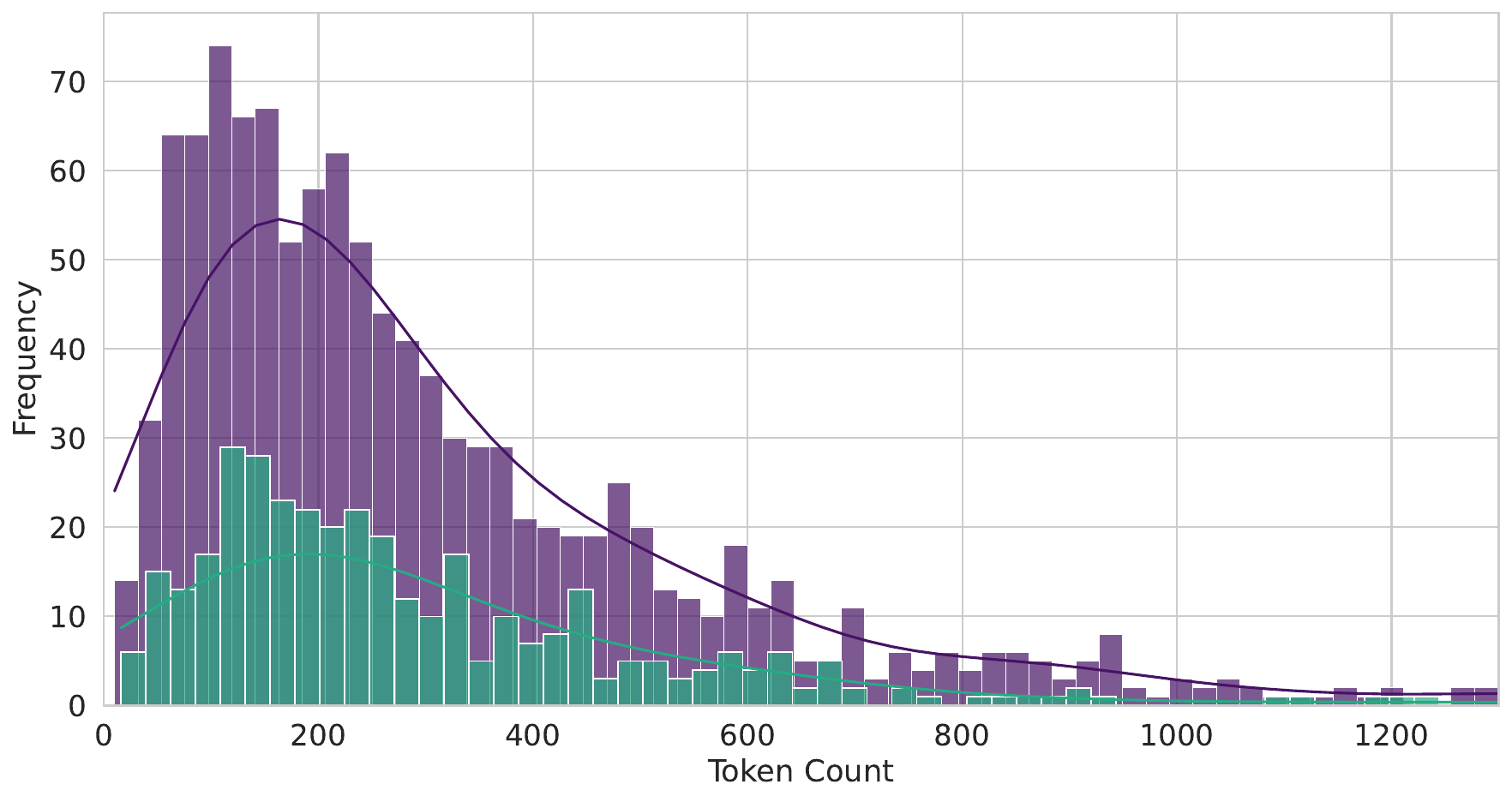}
    \caption{Token distribution for \textbf{BART-large} in user posts. Violet and green represent train and test splits, respectively.}
    \label{fig:tokens-bart-large}
\end{figure*}

\begin{figure*}[h]
    \centering
    \includegraphics[width=0.8\textwidth]{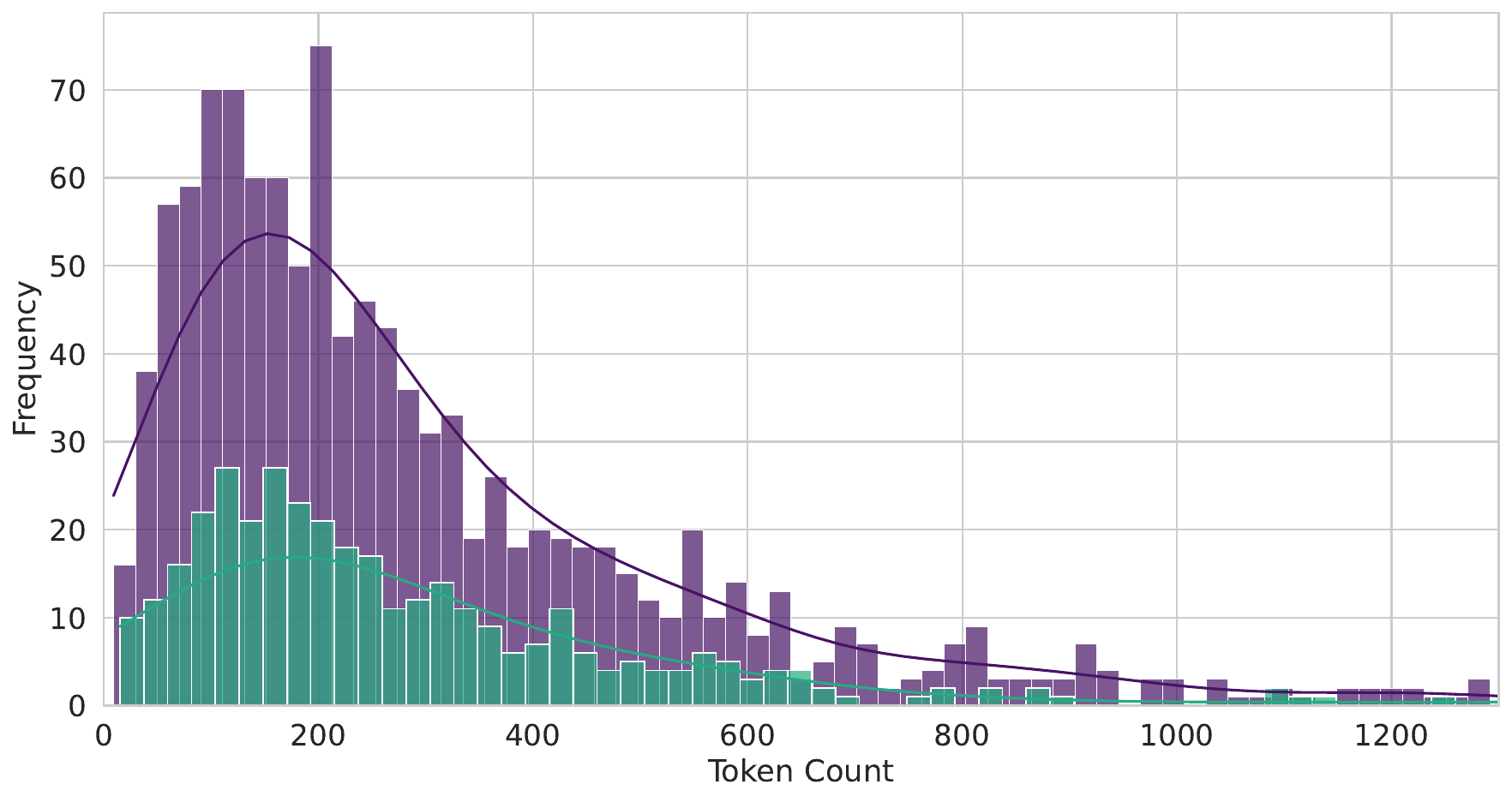}
    \caption{Token distribution for \textbf{Meta-LLaMA-3.1-8B} in user posts. Violet and green represent train and test splits, respectively.}
    \label{fig:tokens-meta-llama}
\end{figure*}
\label{app_tokens}
This section presents examples of token distributions across different models. Each figure corresponds to a specific model, as indicated by the file name, and showcases the token distribution in user posts. The violet and green bars represent the train and test splits, respectively. Table 4 presents the examples of the datasets with mental health disorders, causes, and whether they are seeking help or not.

\end{document}